\documentclass{article}

   \PassOptionsToPackage{numbers, sort&compress}{natbib}
\usepackage[main,final]{neurips_2026}
\usepackage{graphicx}
\usepackage{multirow}

\usepackage[utf8]{inputenc} 
\usepackage[T1]{fontenc}    
\usepackage{hyperref}       
\usepackage{url}            
\usepackage{booktabs}       
\usepackage{enumitem} 

\usepackage{amsfonts}       
\usepackage{nicefrac}       
\usepackage{microtype}      
\usepackage{xcolor}         
\usepackage{amsmath}        
\usepackage{multirow}
\usepackage{wrapfig}
\usepackage{array} 

\usepackage{threeparttable}

\hypersetup{
    colorlinks,
    linkcolor={blue!64!black},
    citecolor={blue!64!black},
    urlcolor={blue!64!black}
}

\definecolor{StimulusOnset}{HTML}{52A754}
\definecolor{movementOnset}{HTML}{E16D3F}
\definecolor{feedbackOnset}{HTML}{A95097}
\newcommand{\ours}{\textsc{BeeMO}}

\newcommand{\codelink}{\href{https://github.com/nerdslab/beemo}{\texttt{https://github.com/nerdslab/beemo}}}

\title{Scaling Neural Motor Decoding via\\ Decoupled Behavioral Pretraining}

\author{%
Divyansha Lachi \\
University of Pennsylvania \\
Philadelphia, PA, USA \\
\texttt{div11@upenn.edu}
\And
Vinam Arora \\
University of Pennsylvania \\
Philadelphia, PA, USA \\
\texttt{vinam@upenn.edu}
\AND
Shivashriganesh P. Mahato \\
University of Pennsylvania \\
Philadelphia, PA, USA \\
\texttt{smahato@upenn.edu}
\And
Alexandre Andre \\
University of Pennsylvania \\
Philadelphia, PA, USA \\
\texttt{aandre1@upenn.edu}
\And
Eva L. Dyer \\
University of Pennsylvania \\
Philadelphia, PA, USA \\
\texttt{dyer1@upenn.edu}
}

\newcommand{\noreffootnote}[1]{\begingroup\renewcommand{\thefootnote}{}\footnote{#1}\addtocounter{footnote}{-1}\endgroup}

\begin{document}

\maketitle

\noreffootnote{%
Reviewed on OpenReview:
\url{https://openreview.net/forum?id=I0HCQLIKMJ}.
}

\begin{abstract}

Recent neural decoding approaches achieve strong performance by training on large collections of paired neural-behavioral recordings, but such datasets are expensive, invasive, and difficult to scale. In contrast, behavioral data is abundant and easy to collect, raising the question of whether neural decoding can benefit from decoupling behavioral and neural representation learning. We introduce \ours, a flexible multi-session framework that enables training with arbitrary mixtures of paired neural-behavioral recordings and unpaired behavioral data. A behavior decoder learns movement dynamics directly from behavioral trajectories, while neural activity is incorporated through cross-attention layers when available, allowing behavioral and neural supervision to scale independently. Across multiple intracortical motor datasets and tasks in nonhuman primates, we show that incorporating unpaired behavioral data improves decoding performance, particularly when transferring to new tasks, and that adding even a small amount of behavioral data from the target task further improves transfer. By decoupling neural and behavioral data, we unlock a new dimension for scaling: unpaired behavioral data, which is substantially cheaper and easier to collect than paired neural recordings, can be used directly to improve decoding performance. We further show that behavior-only pretraining, without any neural pretraining, achieves performance comparable to models pretrained on substantially larger neural datasets. Together, our results suggest that scaling behavioral data offers a practical and cost-effective path toward neural decoding models that generalize across subjects and tasks. Code is available at~\codelink.

\end{abstract}

\section{Introduction}
\label{sec:intro}


Recent work has shown that training on large, diverse neural datasets can improve decoding performance and enable transfer across animals and conditions \cite{azabou2024unified, ye2024neural, zhang2024towards, knight2025unified}. 
However, state-of-the-art neural decoders such as POYO \cite{azabou2024unified} and POSSM \cite{ryoo2025generalizable} are still trained in a tightly coupled manner using datasets containing paired neural activity and behavior. As a result, scaling is achieved primarily by incorporating larger collections of jointly recorded neural-behavioral data \cite{azabou2024unified, ryoo2025generalizable}, where neural representations and behavioral dynamics are learned simultaneously from aligned recordings.

This contrasts sharply with trends in modern multimodal foundation models, where scaling is achieved through decoupled pretraining on asymmetric and unpaired data sources. In vision-language and speech-language models, language models are pretrained independently on massive text corpora, while visual or speech encoders are trained separately and later aligned through conditioning mechanisms or cross-modal objectives \cite{radford2021learning, alayrac2022flamingo, li2023blip, baevski2020wav2vec, radford2023robust}. These decoupled strategies allow each modality to scale according to its own data availability rather than requiring all modalities to be jointly observed at all times.

Neural decoding faces a similar asymmetry. Behavioral data exists at dramatically larger scales than paired neural-behavioral recordings. Across neuroscience, ethology, robotics, and wearable sensing, movement trajectories and behavioral sequences can now be collected continuously and cheaply across diverse conditions \cite{mathis2018deeplabcut, pereira2022sleap}. In many experimental settings, behavioral characterization precedes neural recording by months or years \cite{meister2022learning, oesch2024innate}, producing far larger corpora of behavior than aligned recordings. Yet despite this abundance, most decoding approaches still treat behavior primarily as a supervised target associated with a neural recording, rather than as a modality that can itself be pretrained and scaled independently.

There are strong reasons to expect that decoupling behavioral and neural representation learning can be effective. Motor behavior exhibits rich autonomous temporal structure, where future trajectories are highly constrained by past movement. Classical dynamical-system approaches, including Kalman filtering and latent-state models, explicitly model behavior as an evolving dynamical process and used neural activity primarily to condition or update latent behavioral trajectories \cite{wu2002neural,shenoy2013cortical, yu2007mixture}.

For example, Kalman-filter-based decoders combine a transition model, which captures the expected temporal evolution of behavior, with an observation model that relates neural population activity to behavioral state. This separation allows the decoder to leverage behavioral dynamics while using neural activity to refine or disambiguate the predicted trajectory. These approaches demonstrate that accurate decoding can benefit from explicitly modeling the temporal structure of behavior rather than treating each behavioral output as an independent prediction.



Motivated by these observations, we introduce \ours, a flexible multi-session framework that decouples behavioral and neural representation learning. \ours\ consists of a neural encoder that compresses spike activity into latent neural tokens and a behavior decoder trained to model movement dynamics directly from behavioral trajectories. A behavior-forecasting objective enables the decoder to learn biomechanical regularities and long-range temporal structure independently of neural recordings, while neural embeddings are incorporated through cross-attention layers that condition the behavior model when available. During training, the behavior decoder can be optimized using both paired neural-behavioral recordings and purely behavioral trajectories, allowing behavioral structure to scale independently from neural data availability.

We evaluate \ours\ across multiple intracortical motor decoding tasks in nonhuman primates \cite{dandi:DANDI:000688_0.250122.1735, perich2018neural, ma2023using, gallego2022local}, using two popular neural encoders, POYO \cite{azabou2024unified} and NDT \cite{ye2021representation}, as backbone neural encoders. First, we show that incorporating unpaired behavioral data substantially improves decoding performance, matching the performance of base architectures while using $3\times$ less paired data. Second, we show that behavioral pretraining alone can substitute for substantially larger amounts of neural training data, reducing dependence on jointly recorded datasets. Third, we show that behavioral and neural representations provide complementary sources of information, and that combining both leads to improved prediction of future movements.
Taken together, these findings motivate a shift from tightly coupled decoding pipelines toward modular architectures that independently scale behavior and neural representation learning.

{\bf Our main contributions include:}
\vspace{-3mm}
\begin{itemize}[leftmargin=15pt,itemsep=0em]
\item We propose \ours, a modular framework for neural decoding that decouples behavioral and neural representation learning, enabling behavioral pretraining independently of neural recordings through a conditioning-based architecture.

\item We show that incorporating unpaired behavioral data improves decoding performance across tasks and animals, with \ours\ matching baselines trained on all available paired data using $3\times$ fewer paired neural recording sessions. We further show that incorporating task-aligned behavioral data provides additional gains, improving transfer to the target task.

\item We show that behavior-only pretraining, without any neural pretraining data, outperforms single-session supervised baselines and is competitive with models pretrained on large neural-behavior datasets.



\end{itemize}
\section{Method}
\label{sec:method}

\begin{figure*}[t]
  \centering
  \includegraphics[width=0.95\linewidth]{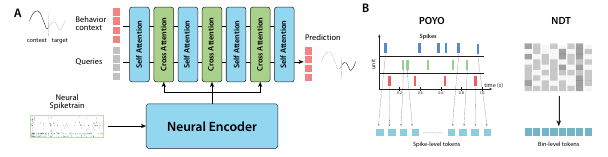}

\caption{\footnotesize{\textbf{Overview of \ours:} \textbf{(A) Architecture:} Behavioral context and query tokens pass through alternating self-attention and cross-attention layers. The behavioral and neural components are decoupled: the behavior decoder operates independently on unpaired behavioral data, and neural latents are incorporated via cross-attention when available.
\textbf{(B) Neural Encoders:}} POYO (left) represents neural activity as spike-level tokens, where each spike is embedded with its timing and unit identity. NDT* (right) represents neural activity as bin-level tokens, aggregating spikes into fixed time bins across units.}
  \label{fig:model_overview}
\end{figure*}

A central challenge in neural decoding is the limited availability of paired neural and behavioral recordings. In contrast, collecting behavioral data is significantly more scalable, as it does not rely on invasive hardware. We leverage this data asymmetry by designing a framework capable of learning from both isolated behavioral trajectories and paired neural-behavioral trajectories. This is enabled by a modular architecture whose behavioral and neural components can be trained independently or jointly, depending on data availability.


Our approach is grounded in two key observations:
\begin{itemize}[leftmargin=15pt,itemsep=0pt,topsep=0pt]
\item \textbf{Behavioral priors:} Motor behavior exhibits strong temporal autocorrelation, making past observations highly predictive of future states. A behavior-forecasting objective exploits this structure, to capture movement dynamics and biomechanical regularities from behavioral data.
\item \textbf{Neural conditioning:} Neural activity conditions the behavioral decoder, providing the neural-specific control signal that resolves ambiguity in the decoded trajectory \cite{pandarinath2018inferring, churchland2012neural} beyond what behavioral context alone can determine.
\end{itemize}

\subsection{Problem Setup}

We consider trajectories of neural activity and behavioral measurements observed over time, where the model takes either or both modalities as input context and predicts behavior as output.
 
\vspace{-3mm}
\paragraph{Neural data:}
Spike-sorted neural recordings are represented as sets of time-stamped spike events across a population 
of neurons. Each observation is a collection of $N_s$ tuples $\{(t_i, n_i)\}_{i=1}^{N_s}$, where 
$t_i$ denotes the spike timestamp and $n_i$ indexes neuron identity. The number of neurons 
and their identities may vary across sessions, with no cross-session neuron alignment assumed.

\vspace{-3mm}
\paragraph{Behavioral data:}
Behavior is represented as a multivariate time series 
$\{b_t\}_{t=1}^{T}$, where
$b_t \in \mathbb{R}^{d_b}$ denotes behavioral variables (e.g., kinematics, cursor
position, or force) at time $t$. Across training regimes, past observations
serve as context, future timesteps serve as prediction targets,
or both.

\vspace{-3mm}
\paragraph{Evaluation protocol:}
At test time, consistent with standard neural decoding benchmarks~\cite{pei2021neural, andre2026brainwidebench}, the model receives 
neural activity over a context window $[0, T_{\text{ctx}}]$ and predicts the corresponding behavior over the same window, with no behavioral context provided as input.

\subsection{Architecture}
\label{sec:arch}

\ours\ is composed of two main components: a neural encoder that compresses spike population activity into a compact set of latent tokens, and a behavior decoder that predicts behavior at target timesteps from available input context. The two components interact via cross-attention, where the behavior decoder optionally conditions on the neural latents. Formally:
\begin{equation}
    \hat{y}_t = \mathrm{\ours}\!\left(H_{\text{ctx}};\, H_{\text{neural}}\right),
\end{equation}
where $H_{\text{ctx}}$ denotes behavioral context tokens, $H_{\text{neural}}$ denotes neural latent tokens, and either input is optional but at least one must be present. This modularity enables the asymmetric training strategy in Section~\ref{sec:training}.

\vspace{-2mm}
\paragraph{Neural encoder.}

The neural encoder maps spike population activity over a context window 
$[0, T_{\text{ctx}}]$ to a fixed set of $M$ latent tokens:
\begin{equation}
    H_{\text{neural}} = \mathrm{NeuralEncoder}\!\left(\{(t_i, n_i)\}_{t_i \in [0, T_{\text{ctx}}]}\right) \in \mathbb{R}^{M \times d}.
\end{equation}

Any encoder that maps neural data to a fixed-size token set can serve in 
this role~\citep{ye2024neural, aroraknow, ryoo2025generalizable}. In this 
work, we consider two neural encoders that differ primarily in their 
tokenization scheme. We refer the reader to Appendix~\ref{app:neural_encoder} for further details.

\begin{itemize}[leftmargin=15pt,itemsep=0pt,topsep=0pt]
\item{\em POYO ~\cite{azabou2024unified}:} operates on raw spike times. Each 
spike $(t_i, n_i)$ is represented by a learnable neuron identity embedding 
with spike timing encoded via RoPE~\cite{su2024roformer} (see 
Appendix~\ref{app:rope} for details). Because the number of spikes is 
variable, POYO uses cross-attention as its first layer, attending from a 
fixed set of $M$ learnable latent tokens over the spike token embeddings 
$S \in \mathbb{R}^{N_s \times d}$, followed by $L$ layers of self-attention 
to produce $H_{\text{neural}} \in \mathbb{R}^{M \times d}$.

\item{\em NDT*~\cite{ye2021representation}:} is a supervised modified variant of the 
Neural Data Transformer that operates on binned spike counts, producing a 
fixed-size sequence of $T_b$ bin tokens $X \in \mathbb{R}^{T_b \times d}$ 
via a learned unit-weighted aggregation scheme (see 
Appendix~\ref{app:neural_encoder} for details). Since the input is 
already a fixed-size sequence, all layers use self-attention. Our modified 
variant incorporates unit stitching~\cite{ye2024neural, zhang2024towards} 
and RoPE~\cite{su2024roformer}, and uses a cross-attention decoder 
following~\cite{azabou2024unified}. We denote this model as NDT* to 
distinguish it from the original.
\end{itemize}

\vspace{-2mm}

\paragraph{Behavior decoder.}
The behavior decoder operates on two sets of tokens: \emph{context tokens} 
derived from past behavioral observations, and \emph{prediction tokens} at 
target timesteps whose values are to be predicted.
Past behavioral observations $\{b_\tau\}_{\tau < t}$ are linearly projected 
into the model dimension to give context tokens 
$H_{\text{ctx}} \in \mathbb{R}^{C \times d}$. Prediction tokens 
$H_{\text{pred}} \in \mathbb{R}^{Q \times d}$ are initialized from a learned 
session embedding at each target timestep. Both sets are augmented with rotary position encodings~\cite{su2024roformer} and concatenated into a single 
sequence of $N = C + Q$ tokens:
\begin{equation}
    H_{\text{behav}}^{(0)} = \bigl[H_{\text{ctx}},\; H_{\text{pred}}\bigr] \in \mathbb{R}^{N \times d}.
\end{equation}
This sequence is processed through $K$ transformer blocks. Within each block, 
self-attention operates over all behavioral tokens, followed by cross-attention 
where the behavioral tokens serve as queries and the neural latents 
$H_{\text{neural}}$ serve as keys and values:
\begin{equation}
    \hat{H}^{(k)} = \mathrm{SelfAttn}\!\left(H_{\text{behav}}^{(k)}\right), H_{\text{behav}}^{(k+1)} = \mathrm{CrossAttn}\!\left(\hat{H}^{(k)},\; H_{\text{neural}}\right)
\end{equation}
\textbf{The cross-attention step is skipped when no neural data is present.} 
After $K$ blocks, a final self-attention layer is applied over all tokens, 
after which the prediction tokens are extracted and passed through a linear 
readout head to produce $\hat{y}_t \in \mathbb{R}^{d_b}$.

\subsection{Training}
\label{sec:training}

The modular design of \ours\ enables a flexible training strategy that exploits 
the data asymmetry described above. All three training regimes, behavior-only, 
joint, and neural-only (via behavior dropout, described below), share the same regression loss:
$\mathcal{L} \;=\; \sum_{t} \left\|\hat{y}_t - y_t\right\|_2^2,
$
where $\hat{y}_t$ is the predicted behavior and $y_t$ is the ground truth.

\paragraph{Asymmetric training.}
Depending on the available input context, the model operates in different modes: when only behavioral data is present, the cross-attention step is skipped; when spike data is available, the behavior decoder additionally conditions on the neural latents via cross-attention. This allows a batch 
to contain a mix of behavior-only and paired sessions, enabling large behavioral datasets to be combined with limited paired recordings within a single training run, with no changes to the loss or architecture.

\vspace{-3mm}
\paragraph{Behavior dropout.} Following prior work and established evaluation protocols, no behavioral context is provided at test time, and the model must rely solely on neural data to reconstruct behavior. Since \ours\ can condition on both neural and behavioral context, this introduces a train-test distribution shift. Without any intervention, the decoder would always have access to behavioral context during training but none at evaluation. To ensure the model can decode from neural activity alone at evaluation, we apply \emph{behavior dropout} during training. With probability $p_{\mathrm{drop}}$, the behavioral context tokens are masked out for paired sessions, forcing the decoder to rely solely on the neural latents and session embeddings to reconstruct behavior. We evaluate the effectiveness of behavior dropout by ablating it in Section~\ref{sec:ablation}.

\section{Results}
\label{sec:results}

We evaluate \ours\ across three settings. First, we limit neural data and examine whether behavioral pretraining can compensate, reducing reliance on large neural recording datasets. Second, we investigate whether the composition of the behavioral pretraining set matters, and whether incorporating task-matched behavioral data improves transfer. Finally, we remove neural pretraining entirely and evaluate how well a behavior-only pretrained model performs when fine-tuned for neural decoding.


\subsection{Experimental setup}

We evaluate on multiple datasets spanning three tasks: center-out reaching (CO) \cite{dandi:DANDI:000688_0.250122.1735, perich2018neural}, random-target reaching (RT) 
\cite{dandi:DANDI:000688_0.250122.1735}, and isometric wrist (IW) 
\cite{ma2023using, ma2023dryad}, covering a range of behavioral complexity (see 
Appendix~\ref{app:dataset} for full dataset details). Evaluation is performed on held-out animals not seen during pretraining, and performance is reported using \(R^2\). Following the evaluation protocol used in~\cite{azabou2024unified, ye2023ndt2}, for each session, we concatenate predictions and ground-truth values across all held-out test trials and compute a single session-level \(R^2\). We then report the unweighted mean across sessions, so that each session contributes equally regardless of the number or duration of its test trials. Unlike prior work, we use a causal train/test split and do not apply trial-based loss weighting, which we believe better reflects realistic deployment settings. As a result, our results are not directly comparable to previously reported numbers, even on the same datasets. All baselines were tuned via extensive hyperparameter search; implementation details and search spaces are provided in Appendix~\ref{app:implementation} and Appendix~\ref{app:hyperparameters}.

We compare against two categories of baselines: (i) single-session fully supervised models (Linear, MLP, GRU, POYO-SS, NDT*-SS) trained on paired neural and behavioral data from a single session, and (ii) multi-session supervised models (POYO, NDT*) trained jointly across sessions.\footnote{We attempted to reproduce NEDS as a baseline using the publicly available code, but were both unable to train it successfully on this data or reproduce their original results.} 
We additionally compare against classical Kalman-filter decoders and POYO with post hoc Kalman smoothing in Appendix~\ref{app:kalman}. The single-session models are highly sensitive to session-specific hyperparameters and therefore undergo extensive tuning independently for each held-out session, making them strong but computationally expensive baselines.\footnote{Please note that all single-session models undergo extensive per-session hyperparameter tuning, with a 100-trial search per held-out session, resulting in 2300 training runs per model across 23 evaluation sessions. Pretrained models, however, use a fixed hyperparameter configuration across sessions. We include the single-session models as strong session-specific reference baselines, with the primary comparisons made against the matched multi-session backbones. See Appendix~\ref{app:hyperparameters} for details.} Our primary comparisons are therefore against the corresponding multi-session backbones. Since \ours\ is modular, we evaluate it with both POYO and NDT* backbones, denoted \ours-POYO and \ours-NDT* respectively. In each case, the only difference from the corresponding multi-session model is access to unpaired behavioral data during pretraining. We additionally ablate \ours\ under different training regimes in Section~\ref{sec:behav_only} and with varying amounts of neural data in Section~\ref{sec:scaling_analsys}.

{\bf Finetuning protocol.} All pretrained models were finetuned using a gradual unfreezing strategy. We first train only newly initialized session-specific parameters (unit embeddings and session embeddings) while keeping the pretrained backbone frozen. After a fixed number of epochs, we unfreeze the backbone and continue training all parameters jointly. (Refer to Appendix~\ref{app:implementation} for details.)

\subsection{Leveraging unpaired behavioral data to improve motor decoding}
\label{sec:boyo_asymmetric}

\begin{table}[t]
\centering
\caption{\footnotesize{\textbf{Asymmetric training results.} All multi-session models are pretrained on center-out (CO) reaching data and evaluated on within-task (CO) and cross-task (RT, Isometric Wrist) transfer. \ours\ variants incorporate $\sim$6.5 hours of additional unpaired behavioral data during pretraining, while base models use paired neural-behavioral data only. Performance is reported as mean $\pm$ SEM over three seeds. \textit{Paired $\Delta$} reports the mean paired difference in $R^2$ between each \ours\ variant and its matched backbone across held-out sessions. Bold indicates a significant improvement ($p<0.05$, two-sided paired $t$-test). Session-wise mean $\pm$ SD and exact $p$-values are reported in Table~\ref{tab:sessionwise_ttest}.}}
\label{tab:boyo_asymmetric}
\resizebox{0.75\textwidth}{!}{
\begin{tabular}{l@{\hspace{1.5cm}}c@{\hspace{1.5cm}}cc}
\toprule
& \textbf{Within-Task} 
& \multicolumn{2}{c}{\textbf{Cross-Task}} \\
\cmidrule{2-2} \cmidrule(lr){3-4}
\textbf{Method} & \textbf{CO} & \textbf{RT} & \textbf{Isometric Wrist} \\
\midrule
\multicolumn{4}{l}{\textit{Single-Session}} \\[2pt]
\quad Linear   & $0.6816 \pm 0.0011$ & $0.6146 \pm 0.0003$ & $0.5826 \pm 0.0006$ \\
\quad MLP      & $0.8995 \pm 0.0014$ & $0.7477 \pm 0.0014$ & $0.7392 \pm 0.0035$ \\
\quad GRU      & $0.9100 \pm 0.0015$ & $0.7839 \pm 0.0016$ & $0.7888 \pm 0.0011$ \\
\quad POYO-SS  & $0.8438 \pm 0.0030$ & $0.7219 \pm 0.0101$ & $0.6967 \pm 0.0063$ \\
\quad NDT*-SS  & $0.8827 \pm 0.0002$ & $0.7693 \pm 0.0007$ & $0.7749 \pm 0.0012$ \\
\midrule
\multicolumn{4}{l}{\textit{Multi-Session (NDT*)}} \\[2pt]
\quad NDT*        & $0.9010 \pm 0.0032$ & $0.7600 \pm 0.0062$ & $0.7767 \pm 0.0017$ \\
\quad \ours-NDT*  & $0.9027 \pm 0.0016$ & $0.7734 \pm 0.0018$ & $0.7802 \pm 0.0001$ \\
\quad\quad Paired $\Delta$ & $+0.0017$ & $\mathbf{+0.0134}$ & $+0.0035$ \\
\midrule
\multicolumn{4}{l}{\textit{Multi-Session (POYO)}} \\[2pt]
\quad POYO        & $0.8973 \pm 0.0013$ & $0.7498 \pm 0.0015$ & $0.7237 \pm 0.0018$ \\
\quad \ours-POYO  & $0.9031 \pm 0.0003$ & $0.7609 \pm 0.0014$ & $0.7361 \pm 0.0020$ \\
\quad\quad Paired $\Delta$ & $+0.0058$ & $\mathbf{+0.0112}$ & $\mathbf{+0.0124}$ \\
\bottomrule
\end{tabular}
}
\vspace{-4mm}
\end{table}


A central challenge in scaling neural decoders is the high cost of obtaining paired neural recordings. We show that performance can be improved by incorporating additional unpaired behavioral data. We compare NDT* and POYO base models against our \ours\ variants, which utilize $\sim$6.5 hours of additional behavior during pretraining. When trained on the CO task, the \ours\ models improve performance in both within-task evaluation on CO and cross-task transfer to RT and IW. As shown in Table~\ref{tab:boyo_asymmetric}, both \ours-POYO and \ours-NDT* outperform their respective multi-session backbones on all tasks, with significant gains in cross-task transfer. \ours-POYO significantly improves over POYO on RT (paired $\Delta R^2=+0.0112$) and IW ($+0.0124$), while \ours-NDT* significantly improves over NDT* on RT ($+0.0134$). The paired improvement for \ours-NDT* on IW is smaller and not significant ($+0.0035$), which we examine further in Section~\ref{sec:task_ablation}. Improvements on the CO task are relatively modest for both models, likely because performance on this within-task evaluation is already near saturation. Session-wise statistics and exact $p$-values are reported in Appendix~\ref{app:sessionwise_stats}. Since \ours\ and the base models share the same underlying architecture, these gains are attributable to the use of unpaired behavioral data during pretraining rather than any architectural advantage. We further confirm this in Section~\ref{sec:ablation}, where replacing POYO’s decoder with \ours’s deeper decoder, without access to additional unpaired behavioral data, does not improve performance.


\subsection{Scaling analysis}
\label{sec:scaling_analsys}

\begin{wrapfigure}{r}{0.33\textwidth}
    \centering
    \vspace{-11mm}
    \includegraphics[width=\linewidth]{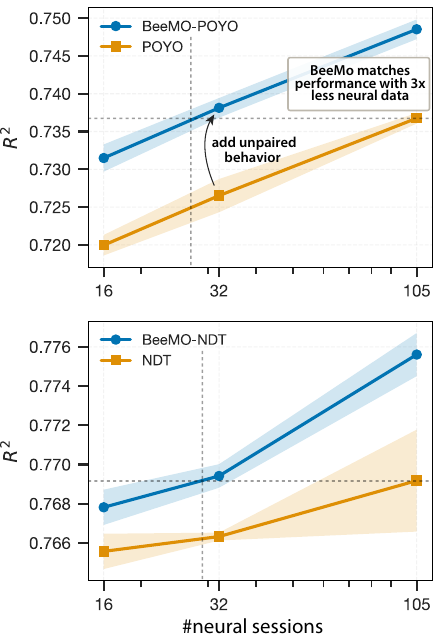}
    \caption{\footnotesize{\textbf{Scaling Analysis:} Average cross-task decoding performance ($R^2$ averaged across RT and IW tasks) as a function of the number of paired neural sessions used during pretraining (models pretrained on CO). Top: BEEMO-POYO versus POYO. Bottom: BEEMO-NDT* versus NDT*. \ours\ outperforms both POYO and NDT* at matched neural data budgets by leveraging unpaired behavioral data during pretraining. Notably, \ours\ matches the performance of POYO and NDT* using approximately 3$\times$ less neural data. Shaded regions denote SEM over three seeds.
    }}
    \label{fig:beemo_scaling}
    \vspace{-4mm}
\end{wrapfigure}

We investigate how \ours\ scales with the amount of available neural data. 
We fix the behavioral pretraining budget for \ours\ to all available CO behavioral data, vary the number of neural sessions, and evaluate cross-task transfer to RT and IW (Figure~\ref{fig:beemo_scaling}). The base performance at each 
operating point is set by the underlying POYO and NDT* backbones, and since 
the behavioral data is held fixed throughout, any gain over these baselines 
reflects how well the cross-attention interface learns to exploit the 
pretrained behavioral representations as a function of neural data. 
Interestingly, the two neural encoders exhibit qualitatively different scaling behavior. \ours-NDT* benefits increasingly from additional neural sessions, with the gap over the NDT* backbone widening as more paired data becomes available, suggesting that its cross-attention layers continue to improve as they are exposed to more neural supervision. \ours-POYO, by contrast, maintains a roughly constant offset over the POYO backbone across all session counts, and does not scale with additional neural data, suggesting the cross-attention interface is unable to further leverage the pretrained behavioral representations as more sessions become available. We believe these differences arise from the distinct tokenization schemes of the two encoders, though a thorough investigation is left for future work. Despite this difference in scaling behavior, both \ours-NDT* and \ours-POYO improve over their respective backbones at every neural data budget.
Notably, \textbf{\ours\ matches the full-data performance of both POYO and NDT* backbones using as few as 32 neural sessions ($\sim$7 hours of paired data), roughly $
\mathbf{3\times}$ less neural data than the full budget of 105 sessions ($\sim$22.5 hours)}. This demonstrates that unpaired behavioral pretraining can effectively substitute for a substantial portion of expensive neural recordings, significantly reducing the data requirements for strong decoding performance.


\subsection{Effect of incorporating task-specific behavioral data}
\label{sec:task_ablation}

\begin{figure*}[t]
  \centering
  
\includegraphics[width=0.95\linewidth]{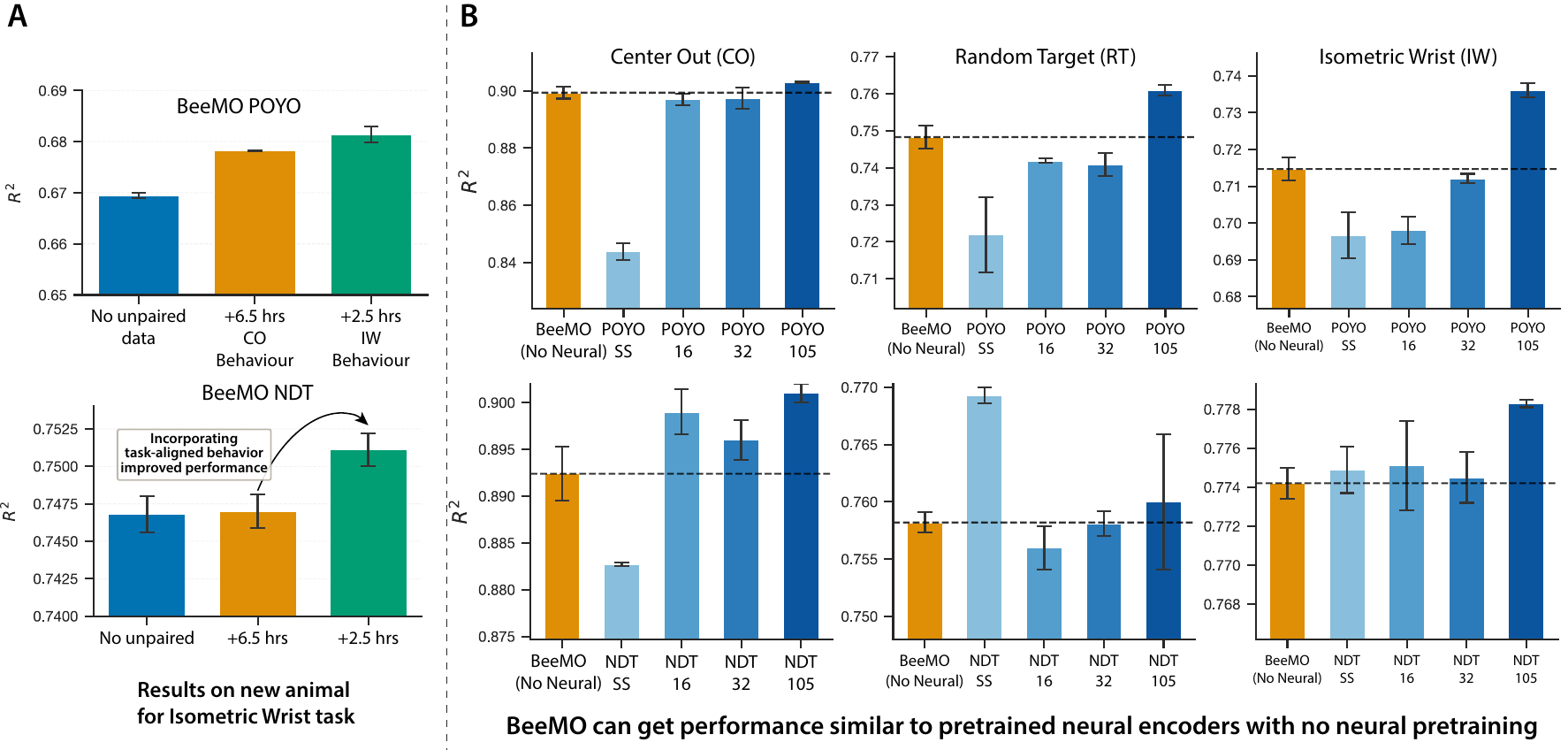}
\caption{\footnotesize \textbf{Behavioral pretraining improves neural decoding.} Error bars denote SEM over three seeds. \textbf{(A) Task-specific behavioral data:} Adding task-matched behavioral data (IW) to the pretraining corpus consistently improves downstream decoding performance on the IW task on a new animal for both \ours-POYO and \ours-NDT*. \textbf{(B) Behavior-only pretraining:} Decoding performance across three tasks (CO, RT, IW) for single-session models and multi-session neural-pretrained models, compared against \ours\ with no neural pretraining (behavior-only).}
  \label{fig:beemo_panel}
  \vspace{-4mm}
\end{figure*}

Having established that unpaired behavioral data improves decoding performance, we next ask whether the composition of that pretraining corpus matters. We hypothesize that behavioral data drawn from the target task distribution carries more signal for downstream transfer than equivalent data from another task which might have different patterns and dynamics. To test this, we augment the pretraining corpus with 2.5 hours of additional IW-task behavioral data from \cite{ma2023using} and evaluate transfer on 7 sessions from a held-out animal. As shown in Figure~\ref{fig:beemo_panel}A, task-matched behavioral pretraining consistently improves downstream decoding for both \ours-POYO and \ours-NDT* (green versus orange), yielding gains of approximately 0.015 and 0.0035 ($R^2$), respectively. This effect is particularly pronounced for \ours-NDT*, where adding 6 hours of CO behavioral data produces no significant improvement on IW task, whereas adding only 2.5 hours of IW behavioral data leads to a substantial performance boost. This suggests that the benefits of behavioral pretraining depend not only on the quantity of behavioral data, but also on its composition and alignment with the downstream task distribution. More broadly, these results indicate that task-specific behavioral data can be collected independently of neural recordings and leveraged during pretraining to improve neural decoding performance.  We observe a similar benefit from adding task-specific behavioral data on RT in Appendix~\ref{app:rt_behavior}.

\subsection{Behavior-only pretraining}
\label{sec:behav_only}

The asymmetric training results raise a natural follow-up question: \emph{how far can behavioral data alone take us in the absence of neural pretraining?} To study this, we pretrain a behavior decoder on large-scale behavioral data without access to neural recordings, and then fine-tune randomly initialized neural encoder and cross-attention layers using paired neural-behavioral data from the target session only. 
As shown in Figure~\ref{fig:beemo_panel}B, behavior-pretrained representations provide a strong inductive bias for neural decoding. Behavior-only \ours-POYO consistently outperforms its single-session baseline despite using the same amount of paired neural data, suggesting that behavioral pretraining alone can meaningfully improve decoding. For \ours-NDT*, even joint neural-behavioral pretraining struggles to outperform the single-session NDT* baseline across most tasks\footnote{Note that all single session models go through extensive per-session hyperparameter tuning, whereas pretrained models use a fixed hyperparameter configuration across all sessions, making them a challenging baseline to surpass for certain neural encoders.}. However, in CO task both joint and behavior-only \ours-NDT* show significant improvements over the single-session baseline, mirroring the pattern observed for POYO. Remarkably, behavior-only \ours-POYO matches and sometimes exceeds POYO pretrained on 32 sessions of joint neural-behavioral data (7 hours of CO neural recordings), despite using no neural pretraining data at all. In general behavior-only \ours\ get competitive performance without access to any neural pretraining. These findings suggest that a substantial portion of the gains commonly attributed to neural pretraining can instead emerge from behavior-only pretraining, offering a more scalable path for neural decoding.

\subsection{How much does providing past behavior improve decoding?}
\label{sec:traj}

\begin{figure*}[t]
  \centering
\includegraphics[width=\linewidth]{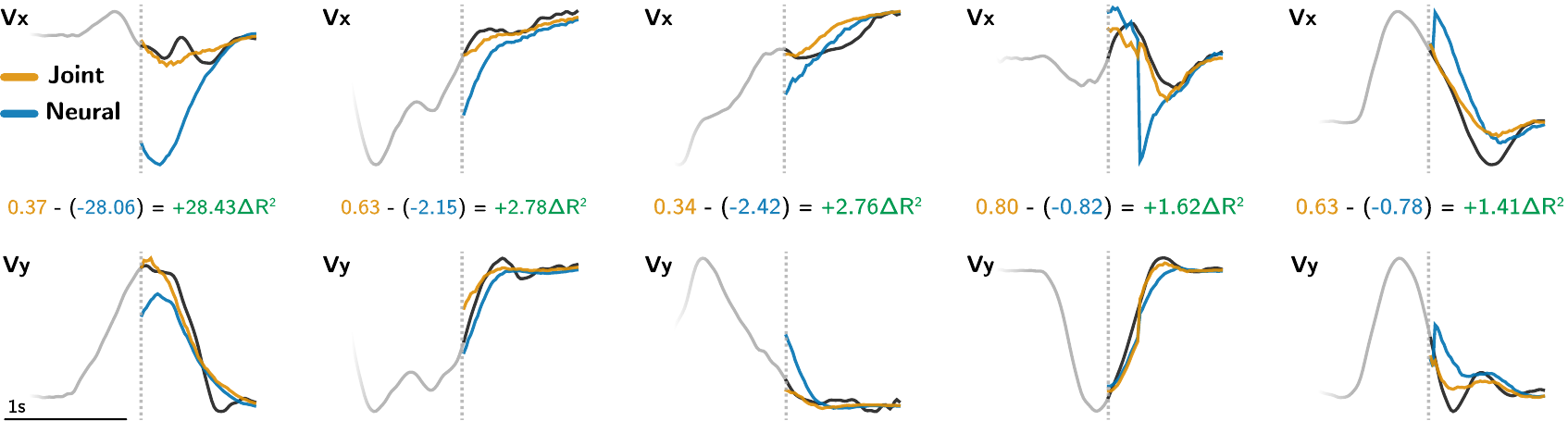}
\caption{\footnotesize \textbf{Behavioral context improves neural decoding.} 
Predicted velocity trajectories ($v_x$, $v_y$) under neural-only (blue) and 
joint neural-behavior (orange) context, vs. ground truth (black), 
across 1~sec decoding windows sorted by $\Delta R^2$ 
(largest to smallest improvement).  
Additional trajectory examples are provided in Figure~\ref{fig:additional_trajectories} in the Appendix.
}
  \label{fig:traj_plots}
  \vspace{-5mm}
\end{figure*}

Behavior exhibits strong temporal continuity, motivating the question of whether past behavioral context can improve decoding beyond neural activity alone. To study this, we compare neural-only decoding with joint neural-behavior decoding. As shown in Figure~\ref{fig:traj_plots}, neural-only context captures the general timing and direction of movements but often produces jerky and poorly scaled trajectories, leading to large amplitude and smoothness errors. This effect is particularly pronounced near the beginning of the decoding window, where the model has little preceding neural context. As the window progresses, more neural history becomes available and neural-only predictions improve, eventually approaching the joint model. 

In contrast, past behavioral context (preceding 1 sec) immediately grounds the prediction in the current behavioral state, producing substantially smoother and more coherent trajectories with improved amplitude calibration and more accurate trajectory initialization. Predicted onsets align more closely with the target signal from the start of the decoding window, suggesting that behavioral context provides a strong dynamical prior over plausible kinematic trajectories. 

\subsection{Ablations}
\label{sec:ablation}

To isolate the contribution of our asymmetric training strategy from the minor architectural modifications introduced in \ours, we conduct two targeted ablations.

\begin{wraptable}{r}{0.5\textwidth}
\centering
\vspace{-6.5mm}
\caption{\footnotesize{\textbf{Ablation study:}
Values are reported as mean $\pm$ SEM across three seeds.}}
\label{tab:ablations}
\vspace{1mm}
\resizebox{0.48\textwidth}{!}{
\begin{tabular}{l|cc}
\toprule
& \multicolumn{2}{c}{\textbf{Cross-Task}} \\
\cmidrule(lr){2-3}
\textbf{Model Variant} & \textbf{RT} & \textbf{IW} \\
\midrule
\textbf{\ours-POYO}
& $0.7609{\scriptscriptstyle\,\pm\,0.0014}$
& $0.7361{\scriptscriptstyle\,\pm\,0.0020}$ \\

\quad w/o dropout
& $0.6925{\scriptscriptstyle\,\pm\,0.0133}$
& $0.6898{\scriptscriptstyle\,\pm\,0.0035}$ \\

\quad w/o unpaired
& $0.7476{\scriptscriptstyle\,\pm\,0.0012}$
& $0.6849{\scriptscriptstyle\,\pm\,0.0017}$ \\
\midrule

\textbf{\ours-NDT*}
& $0.7734{\scriptscriptstyle\,\pm\,0.0018}$
& $0.7802{\scriptscriptstyle\,\pm\,0.0001}$ \\

\quad w/o dropout
& $0.7101{\scriptscriptstyle\,\pm\,0.0083}$
& $0.7266{\scriptscriptstyle\,\pm\,0.0019}$ \\

\quad w/o unpaired
& $0.7683{\scriptscriptstyle\,\pm\,0.0030}$
& $0.7757{\scriptscriptstyle\,\pm\,0.0011}$ \\
\bottomrule
\end{tabular}
}
\vspace{-3mm}
\end{wraptable}

\vspace{-2mm}
\paragraph{Behavior dropout.} At test time, no behavioral context is provided, making behavior dropout critical for \ours\ to perform well under standard evaluation protocols. We verify this by training \ours\ without behavior dropout and find that performance drops significantly for both variants, with \ours-POYO and \ours-NDT* degrading by roughly 6--9\% $R^2$ across tasks (Table~\ref{tab:ablations}). Notably, these degraded models fall below even their respective base model performances, with \ours-POYO and \ours-NDT* dropping by 0.0573 and 0.0499 $R^2$ below POYO and NDT* on RT respectively (Refer Table~\ref{tab:ablations} and ~\ref{tab:boyo_asymmetric}). This confirms that without behavior dropout, the additional unpaired behavioral data not only fails to help but actively hurts performance due to the distribution shift between train and test.

\vspace{-2mm}
\paragraph{Decoder capacity.}
\ours\ uses a deeper behavior decoder than the base neural encoders, consisting of alternating self-attention and cross-attention layers, compared to the single cross-attention decoder used in both POYO and NDT*. This difference in capacity could in principle account for some of the observed performance gains. To disentangle decoder capacity from the effect of unpaired behavioral data, we replace the default decoder in both POYO and NDT* with \ours's alternating self-attention and cross-attention decoder while keeping the original training regime with no access to unpaired behavioral data. As shown in Table~\ref{tab:ablations}, this model does not recover \ours's performance and in fact leads to a slight degradation relative to the base models, most notably for the POYO variant which drops by 0.0388 $R^2$ on IW, likely due to the increased model capacity without a corresponding increase in training data. This demonstrates that the performance gains of \ours\ stem from the additional unpaired data rather than the increased decoder capacity.

\section{Related Work}
\label{sec:relatedwork}

{\bf Neural decoding and its applications.}
Neural decoding aims to recover behavior, sensory variables, or cognitive states from neural activity \cite{paninski2007statistical,glaser2020machine}. This has implications in both our understanding of where different signals or information is processed and how, and also in brain computer interfaces for assisting users in different tasks by leveraging their brain activity.

Classical methods for the neural decoding of movement ranged from population-vector and linear estimators to Wiener filters and probabilistic state-space models, including Kalman and point-process filters \cite{georgopoulos1986neuronal, brown1998statistical, wu2002neural, eden2004dynamic, kulkarni2008state}. Early nonlinear extensions used artificial and recurrent neural networks \cite{wessberg2000real, sanchez2005interpreting}, while more recent work has adopted increasingly flexible machine-learning architectures \cite{glaser2020machine, livezey2021deep}.
 More recent methods seek representations that generalize across recording sessions or subjects by learning shared neural dynamics \cite{pandarinath2018inferring, sani2024dissociative, vermani2024meta}, using distribution or contrastive alignment \cite{lee2019hierarchical, liu2021drop, schneider2023learnable, karpowicz2025stabilizing, zhang2024exploiting}, or through pretraining on large collections of neural recordings \cite{azabou2023unified,ye2021representation,zhang2024towards, zhangneural, azabou2025multi}. Despite differing architectures, these approaches generally build decoders through paired neural-behavioral observations and do not leverage unpaired behavioral data for pretraining.

{\bf Behavioral dynamics as a prior for neural decoding.}
Using temporal structure in behavior to constrain neural decoding has a long history, particularly in motor brain--computer interfaces. Kalman decoders combine a neural observation model with an explicit state-space model of behavioral dynamics, producing trajectories that are consistent both with the observed neural activity and with assumptions such as smooth, approximately linear evolution of position and velocity \cite{wu2002neural}. This framework has been extended in a number of ways including the use of switching dynamics \cite{wu2004modeling} and feedback-intention recalibration \cite{gilja2012high}. These methods demonstrate the value of a behavioral prior, but typically specify a low-dimensional dynamical model for a particular task and output space.

Behavioral structure has also been exploited at the distributional level to align neural activity across days or animals \cite{dyer2017cryptography, lee2019hierarchical, schneider2023learnable, karpowicz2024few}. 
Such approaches can be effective when the sessions are collected during closely matched tasks, such as stereotyped center-out reaches or maze tasks, but depend heavily on correspondence between the behavioral distributions being aligned. \ours\ instead learns behavioral dynamics directly from diverse, unpaired trajectories and uses paired recordings only to map neural activity into the learned behavioral space. This replaces a fixed task-specific dynamical prior with a reusable, data-driven representation of behavior. 

{\bf Self-supervised representation learning for behavior.}
One line of work seeks to discover recurring structure in behavior: MoSeq \cite{wiltschko2015mapping} and B-SOiD \cite{hsu2021b} segment continuous behavior into discrete motifs, while SKIL \cite{wang2025learning} models behavior as continuous compositions of reusable motor primitives. Complementary approaches learn continuous behavioral representations, using generative or reconstruction-based objectives in BehaveNet \cite{batty2019behavenet} and VAME \cite{luxem2022identifying}, and self-supervised objectives in BAMS \cite{azabou2023relax}  and BEAST \cite{wang2026animal}.  These studies demonstrate that unlabeled behavioral data contain rich temporal structure that can support downstream analysis, but do not link behavior to neural recordings.

{\bf Non-neural pretrained models as decoding priors.}
Related BCI approaches use models trained on abundant non-neural data to constrain decoding from scarce paired recordings. Speech BCIs incorporate language models to rerank or correct decoded sequences \cite{metzger2022generalizable, willett2023high, willett2024brain}, or map neural activity into representations learned by pretrained audio and speech models \cite{defossez2023decoding,zhang2026crossspecies}. Similarly, visual decoding methods map fMRI activity into the latent or conditioning spaces of pretrained image models such as Stable Diffusion and CLIP \cite{takagi2023highresolution,ozcelik2023natural}. \ours~ follows the broader principle of separating large-scale non-neural learning from neural decoding. Rather than using an existing linguistic or perceptual model as an output prior, however, it learns the behavioral representation itself from unpaired movement data and then aligns neural activity to that space.

\section{Conclusion}
\label{sec:conclusion}

In this paper, we introduced \ours, a modular pretraining framework that leverages unpaired behavioral data to improve neural decoding across diverse tasks and animals. Its modular design allows any neural encoder to benefit from behavioral pretraining without architectural modifications. Across multiple datasets and decoding tasks, we show that behavioral pretraining provides a strong inductive bias for neural decoding and that unpaired behavioral data can partially substitute for expensive paired recordings. \ours\ matches cross-task performance of fully trained neural decoders using 3x less neural data and more unpaired behavior data, enabling competitive performance with substantially fewer neural sessions. 


A central property of \ours\ is the decoupling of behavioral and neural representation learning. The behavior decoder is pretrained independently and interfaces with neural encoders through a fixed-size latent bottleneck, allowing improvements in neural encoders to transfer directly without retraining the behavior model. This design enables flexible integration with different encoder architectures and isolates the contribution of behavioral pretraining from architectural changes. In our experiments, \ours\ improves over matched POYO and NDT baselines despite using the same underlying neural encoder, indicating that the gains arise from behavioral pretraining rather than encoder capacity.




More broadly, our results indicate that substantial components of decoding performance can be learned independently of neural recordings: a decoder pretrained entirely on behavior achieves performance comparable to models pretrained on substantially larger neural datasets after limited finetuning. This asymmetry mirrors neuroscience itself, where behavioral characterization often precedes large-scale neural recording by months or years, resulting in far larger corpora of behavioral observations than paired neural-behavioral datasets. As neural encoders continue to improve in cross-subject generalization~\cite{aroraknow}, behavior-pretrained decoders provide a complementary path toward more transferable neural decoding systems.

{\bf Limitations and future work.}
Our study primarily considers motor decoding, where behavioral targets exhibit strong temporal structure. An important next step is to test whether the benefits of behavioral pretraining extend to broader settings, including recent benchmarks spanning brain-wide and multi-region recordings~\cite{andre2026brainwidebench}. In addition, \ours\ requires an observable temporal target or meaningful behavioral proxy, and therefore does not directly apply to cognitive states without such measurements. Our experiments are also largely based on structured, task-driven datasets. Extending \ours\ to unconstrained, freely moving behavior may require richer behavioral representations that capture more heterogeneous and hierarchical structure. \ours\ also incurs greater training cost than simple post hoc dynamics models such as Kalman smoothing, although these models are less flexible across tasks and datasets with different behavioral dynamics or sampling rates (Appendix~\ref{app:kalman}). More broadly, we view \ours\ as an initial demonstration of how large-scale behavioral pretraining can support neural decoding. Recent work has explored learning general-purpose representations of behavior from large unlabeled datasets~\cite{sun2024video, ye2024superanimal, azabou2023relax}, and our results suggest that such representations can provide useful priors for decoding even in the absence of paired neural data. Looking forward, we expect the utility of behavioral pretraining to increase as both behavioral foundation models and transferable neural encoders continue to improve.

\section*{Acknowledgments and Disclosure of Funding}
We thank Konrad Körding, Mehdi Azabou and Keertika Saroj for insightful discussions and feedback on
this manuscript. This project was supported by NSF CAREER Award RI:2146072, NIH grant 1RF1MH140429, as well as generous gifts from the CIFAR Learning in Machines \& Brains Program. This work would not have been possible without the computing resources provided by the Penn Advanced Research Computing Center (PARCC) at the University of Pennsylvania. 
This work also used resources available through the National Research Platform (NRP) \cite{10.1145/3708035.3736060} at the University of California, San Diego. NRP has been developed, and is supported in part, by funding from National Science Foundation, from awards 1730158, 1540112, 1541349, 1826967, 2112167, 2100237, and 2120019, as well as additional funding from community partners.



\bibliographystyle{unsrtnat} 
\bibliography{paper}

\newpage

\appendix
\newpage
\setcounter{figure}{0}
\setcounter{table}{0}
\renewcommand{\thefigure}{A\arabic{figure}}
\renewcommand{\thetable}{A\arabic{table}}

\section*{Appendix}

\section{Additional Model Details}
\subsection{Neural Encoder Details}
\label{app:neural_encoder}

We consider two neural encoder architectures: POYO~\cite{azabou2024unified}, which operates on individual spike events, and NDT, which operates on binned spike counts. For a fair comparison, we adapt the original NDT architecture by incorporating learnable unit embeddings with unit stitching~\cite{ye2024neural, zhang2024towards} and rotary position encodings (RoPE)~\cite{su2024roformer}, neither of which was present in the original NDT. We refer to this modified architecture as NDT*. Below, we describe the tokenization and encoding schemes for each architecture, followed by the shared decoding mechanism.

\subsubsection{POYO}

\paragraph{Spike tokenization.}
We adopt the tokenization scheme from POYO~\cite{azabou2024unified}, where 
each spike is represented using two pieces of information: the identity of 
the neural unit it came from, and the timestamp at which it occurred. The 
former corresponds to a learnable unit embedding for each neuron, while the 
latter is encoded with rotary position encodings 
(RoPE)~\cite{su2024roformer}, which allows tokens to be processed based on 
their relative timing rather than absolute timestamps. Formally, a spike 
from neural unit $i$ at time $t_{\text{spike}}$ is represented as a 
$d$-dimensional token:
\begin{equation}
    \mathbf{x} = \left(\mathrm{UnitEmb}(i),\, t_{\text{spike}}\right),
\end{equation}
where $\mathrm{UnitEmb} : \mathbb{Z} \rightarrow \mathbb{R}^d$ is the unit 
embedding map. A time chunk with $N_s$ spikes produces $N_s$ spike tokens, 
yielding $S \in \mathbb{R}^{N_s \times d}$. We use the term \emph{neural 
unit} rather than neuron, as spikes may be assigned at varying levels of 
specificity depending on the dataset.

\paragraph{Latent encoder.}
POYO maintains a set of $M$ learnable latent tokens 
$H_{\text{neural}}^{(0)} \in \mathbb{R}^{M \times d}$, independent of the 
input spike sequence. These latents are first updated via cross-attention 
over the spike token embeddings $S$:
\begin{equation}
    H_{\text{neural}}^{(0)} \leftarrow H_{\text{neural}}^{(0)} + 
    \mathrm{CrossAttn}\!\left(H_{\text{neural}}^{(0)},\, S\right),
\end{equation}
and then processed through $L$ layers of self-attention:
\begin{equation}
    H_{\text{neural}}^{(l+1)} = H_{\text{neural}}^{(l)} + 
    \mathrm{SelfAttn}\!\left(H_{\text{neural}}^{(l)}\right), 
    \quad l = 0, \ldots, L-1,
\end{equation}
producing the final neural representation 
$H_{\text{neural}} = H_{\text{neural}}^{(L)} \in \mathbb{R}^{M \times d}$. 
The fixed-size output is independent of spike count and neuron identity, 
making it compatible with any downstream decoder.

\subsubsection{NDT*}

\paragraph{Bin tokenization with unit stiching}
NDT* operates on binned spike counts rather than raw spike times. Neural 
activity is divided into $T_b$ fixed-size time bins of width $\Delta t$. 
For each bin $t$, a bin token is formed by aggregating the contributions 
of all recorded units via a weighted sum of learnable unit embeddings:
\begin{equation}
    \mathbf{x}_t = \sum_{n=1}^{N} b_{t,n} \cdot 
    \mathrm{UnitEmb}(n) \in \mathbb{R}^{d},
\end{equation}
where $b_{t,n}$ is the spike count for unit $n$ in bin $t$. This produces 
a fixed-size sequence $X \in \mathbb{R}^{T_b \times d}$, to which RoPE 
is applied using the bin center timestamps. To handle varying neuron 
identities across sessions, we adopt the unit stitching approach 
of~\cite{ye2024neural, zhang2024towards}, which maps units from unseen 
sessions into the learned unit embedding space.

\paragraph{Latent encoder.}
Since the input $X$ is already a fixed-size sequence, NDT* processes it 
through $L$ layers of self-attention:
\begin{equation}
    H_{\text{neural}}^{(l+1)} = H_{\text{neural}}^{(l)} + 
    \mathrm{SelfAttn}\!\left(H_{\text{neural}}^{(l)}\right), 
    \quad l = 0, \ldots, L-1,
\end{equation}
with $H_{\text{neural}}^{(0)} = X$, producing the final representation 
$H_{\text{neural}} = H_{\text{neural}}^{(L)} \in \mathbb{R}^{T_b \times d}$. 
Unlike POYO, no cross-attention is required as the token count is fixed 
by the binning scheme, with $M = T_b$.

\paragraph{Cross-attention decoder.}
The decoder predicts behavior at target timestamps by attending over the 
neural latent tokens $H_{\text{neural}} \in \mathbb{R}^{M \times d}$. 
For each target timestep $t$, a query token is formed from a learnable 
session embedding modulated by the target timestamp:
\begin{equation}
    \mathbf{q}_t = \mathrm{SessionEmb}(s, t) \in \mathbb{R}^{d},
\end{equation}
where $s$ indexes the current session. The query tokens 
$Q \in \mathbb{R}^{T_{\text{out}} \times d}$ then attend over the neural 
latents via cross-attention:
\begin{equation}
    H_{\text{dec}} = \mathrm{CrossAttn}(Q,\, H_{\text{neural}}) 
    \in \mathbb{R}^{T_{\text{out}} \times d},
\end{equation}
and a linear projection maps each output token to the predicted behavioral 
variable:
\begin{equation}
    \hat{y}_t = W_{\text{out}}\, H_{\text{dec},t} \in \mathbb{R}^{d_b}.
\end{equation}
This design is shared across both POYO and NDT*, as the fixed-size latent 
output of both encoders is compatible with the same decoder architecture.

\subsection{Rotary position encodings and value rotation.}
\label{app:rope}
RoPE is applied across all attention layers by incorporating a rotation 
matrix $R(t) \in \mathbb{R}^{d \times d}$ into the attention mechanism, 
replacing the standard attention score with:
\begin{equation}
    a_{ij} = \mathrm{softmax}\!\left((R(t_i)\mathbf{q}_i)^\top 
    (R(t_j)\mathbf{k}_j)\right),
\end{equation}
which depends only on the relative timing $t_j - t_i$ between tokens, 
since $R(t_i)^\top R(t_j) = R(t_j - t_i)$. In addition to the standard 
RoPE formulation, POYO applies rotation to the value vectors as well, 
explicitly encoding relative timing information in the attention output:
\begin{equation}
    \mathbf{z}_j \leftarrow \mathbf{z}_j + R(-t_j) 
    \sum_i a_{ij}\, R(t_i)\mathbf{v}_i,
\end{equation}
where $R(t_i)$ is first applied to each value before the weighted sum, 
and $R(-t_j)$ is applied to the output. This captures relative timing 
information in the output of each attention block, which is not possible 
with attention scores alone.

\section{Training Details}
\label{app:implementation}

\paragraph{Architecture.}
All models use an embedding dimension of $d = 128$ with 8 attention heads and a feed-forward multiplier of 4, with attention dropout of 0.2 and linear dropout of 0.4. The context window is fixed to 1s. For POYO, the neural encoder uses 8 transformer layers with 32 latent tokens per timestep spaced at 0.125s intervals, giving $M = 256$ latent tokens in total. For NDT*, the neural encoder uses 8 transformer layers with a bin size of 20ms, giving $T_b = 50$ bin tokens. The behavior decoder is 
shared across both encoder variants and consists of 4 alternating 
self-attention and cross-attention layers.

\paragraph{Optimization.}
All models are trained using AdamW with a base learning rate of $5 \times 10^{-5}$, scaled linearly with batch size following the linear scaling rule, giving an effective learning rate of $5 \times 10^{-5} \times 256$. Weight decay is set to $10^{-2}$. Models are trained with a linear warmup followed by cosine decay learning rate schedule for 100 epochs with a batch size of 256 in bfloat16 precision.

\paragraph{Data augmentation.}
During training, we apply unit dropout to the neural population, 
sampling between 30 and 300 units per batch with a mode of 100 units, following a triangular distribution with peak weight of 4.
For \ours\, we additionally apply behavior dropout (Section~\ref{sec:training}), randomly masking behavioral inputs for samples where paired neural activity is present, with a dropout probability of 0.5.

\paragraph{Finetuning.}
We finetune all pretrained models using gradual unfreezing. For the first 20 epochs, only the newly initialized session-specific parameters (unit embeddings and session embeddings) are trained while the pretrained backbone remains frozen. The backbone is then unfrozen and all parameters are trained jointly, with early stopping patience of 50 epochs, up to a maximum of 300 epochs total.

\paragraph{Hardware.}
All models are trained on 2 NVIDIA RTX B200 GPUs. Finetuning was done on 1 NVIDIA RTX 5090 GPUs.

\paragraph{Runtime}
Pretraining \ours\ and the corresponding base architectures requires approximately 1-3 hours, depending on the amount of pretraining data used. Fine-tuning a pretrained model across all evaluation sessions requires an additional \(\sim\)2 hours.
\section{Dataset Details}
\label{app:dataset}

For all datasets, we adopt a causal trial-based split. In the standard 
setting, we use the first 60\% of trials for training, the following 20\% 
for validation, and the final 20\% for testing, preserving temporal ordering to prevent data leakage. Prior work such as 
POYO~\cite{azabou2024unified} uses trial structure to apply differential 
loss weighting across task epochs (e.g., hold, reach, and return periods), 
which can improve performance but relies on privileged information that is 
not always available in practice. We remove this assumption and apply 
uniform loss weighting across all timepoints, which we believe better 
reflects a realistic deployment setting.

\begin{figure*}[t]
  \centering
  \includegraphics[width=0.7\linewidth]{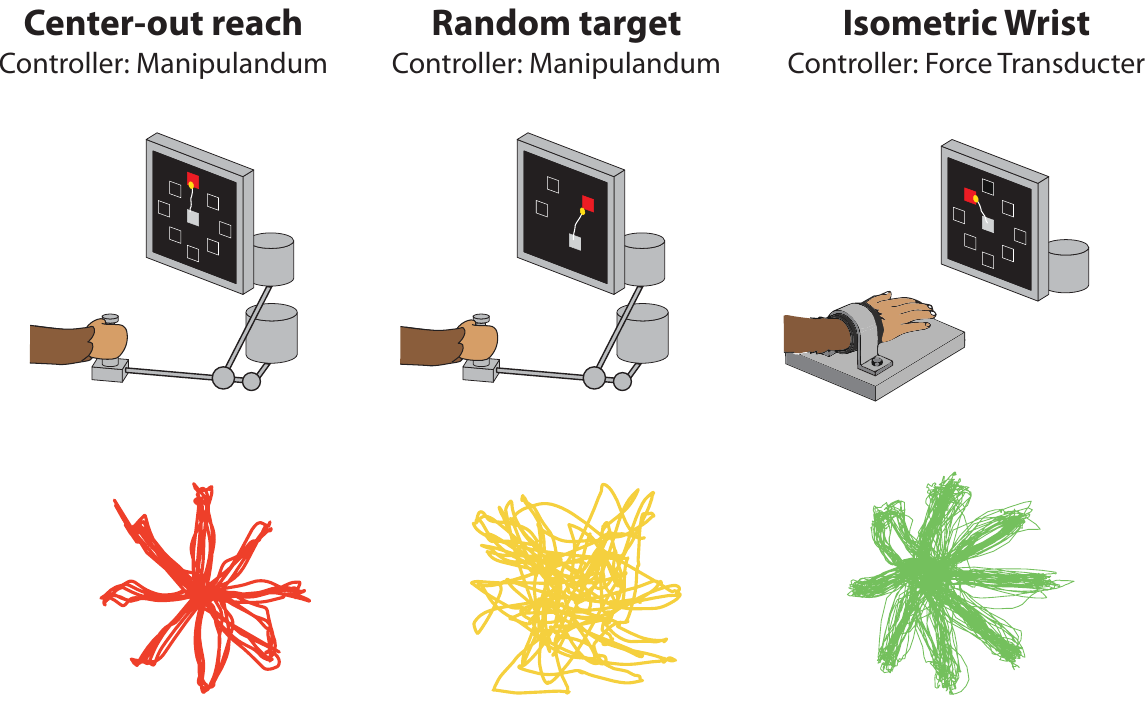}
\caption{\footnotesize{\textbf{Motor tasks and representative behavioral trajectories.} Center-out reaching (CO) and random-target reaching (RT) are performed with a planar manipulandum; isometric wrist (IW) uses a force transducer with no arm movement.}}
  \label{fig:task_diagram}
\end{figure*}

\paragraph{Center-out reaching (Figure~\ref{fig:task_diagram}, left).}
We use recordings from \cite{dandi:DANDI:000688_0.250122.1735, perich2018neural} and 
\cite{gallego2022local, gallego2022dryad}, in which a non-human primate performs reaches to 
targets arranged radially around a central start position. The decoded 
behavioral variable is hand velocity in 2D Cartesian coordinates.

\paragraph{Random-target reaching (Figure~\ref{fig:task_diagram}, middle).}
We use recordings from \cite{dandi:DANDI:000688_0.250122.1735}, in which 
the animal reaches to targets placed at randomized locations. This task 
demands more varied motor planning than center-out reaching, as targets 
are not constrained to fixed directions.

\paragraph{Isometric wrist (Figure~\ref{fig:task_diagram}, right).}
We use recordings from \cite{ma2023using, ma2023dryad}, in which the animal controls 
a cursor by exerting torque at the wrist with no arm movement. The decoded 
behavioral variable is cursor velocity in 2D.

\subsection{Evaluation Sessions}
All evaluations are performed on held-out animals not seen during 
pretraining. For center-out reaching and random-target reaching, we use 
6 sessions each from monkey T from \cite{dandi:DANDI:000688_0.250122.1735}. 
For the isometric wrist task, we use the last 4 sessions from monkey J 
and the last 7 sessions from monkey S from \cite{ma2023using}.

\subsection{Pretraining Sessions}
All center-out sessions from \cite{dandi:DANDI:000688_0.250122.1735}, 
\cite{gallego2022local}, and \cite{ma2023using} not held out for evaluation 
were included in the base pretraining set. Neural spiking data was drawn 
exclusively from \cite{dandi:DANDI:000688_0.250122.1735} and 
\cite{ma2023using}, comprising 105 sessions used for neural pretraining. 
As \cite{gallego2022local} provides only preprocessed binned spike counts 
rather than the raw spikes required by POYO's neural tokenization pipeline, 
only behavioral signals from this source were used. The first 15 isometric 
wrist sessions from monkey J of \cite{ma2023using} are additionally used 
as behavioral pretraining data in the ablation study of 
Section~\ref{sec:task_ablation}.
\section{Hyperparameter Tuning}
\label{app:hyperparameters}

In this section we detail the hyperparameter search configurations for all baseline models. All models are tuned on individual sessions and tasks using the Tree-structured Parzen Estimator (TPE) 
algorithm~\cite{bergstra2011tpe} implemented in 
Optuna~\cite{akiba2019optuna}, with 100 trials per model. For each trial, models are optimized using AdamW with a linear warmup followed by cosine decay learning rate schedule.

\subsection{Linear}

We implement a linear baseline on binned spikes. The model accepts the full 1s input context, bins spike trains at a fixed bin size, and maps directly to the decoding target via a linear readout.
We detail the tuning parameters for linear decoding baseline in Table~\ref{tab:hp_linear}.

\begin{table}[h]
\caption{Hyperparameter search space for the linear baseline, tuned 
for 100 trials.}
\label{tab:hp_linear}
\begin{center}
\begin{tabular}{l l l}
\toprule
\textbf{Hyperparameter} & \textbf{Search Space} & \textbf{Sampling} \\
\midrule
\multicolumn{3}{l}{\textit{Model}} \\[2pt]
\quad Bin size (ms) & $\{1, 5, 10, 20, 40\}$ & Categorical \\[6pt]
\multicolumn{3}{l}{\textit{Training}} \\[2pt]
\quad Number of epochs & $\{100, 300, 500\}$ & Categorical \\
\quad Batch size & $[16, 128]$ & Discrete ($\log_2$) \\
\quad Weight decay & $[10^{-6}, 10^{-1}]$ & Loguniform \\
\quad Base learning rate & $[10^{-5}, 10^{-2}]$ & Loguniform \\
\bottomrule
\end{tabular}
\end{center}
\end{table}

\subsection{Multi-Layer Perceptron (MLP)}

We parameterize our MLP baseline based on network depth and the width of the first hidden layer. Successive hidden layers project inputs down by a factor of $2$, followed by a final readout layer. In initial experiments we found this was a reasonable design choice to balance performance enabled by expressivity of the search space with combinatorial complexity (e.g.,~an alternative is to allow Optuna to select a width for every layer, but the search space becomes much larger). Like the linear baseline, the MLP accepts the full context window at once and bins at a fixed bin size. We detail the tuning parameters for the MLP decoding baseline in Table~\ref{tab:hp_mlp}.

\begin{table}[h]
\caption{Hyperparameter search space for the MLP baseline, tuned 
for 100 trials.}
\label{tab:hp_mlp}
\begin{center}
\begin{tabular}{l l l}
\toprule
\textbf{Hyperparameter} & \textbf{Search Space} & \textbf{Sampling} \\
\midrule
\multicolumn{3}{l}{\textit{Model}} \\[2pt]
\quad Bin size (ms) & $\{5, 10, 20, 40\}$ & Categorical \\
\quad Depth & $\{1, 2, 3\}$ & Categorical \\
\quad Hidden dim & $[32, 512]$ & Discrete ($\log_2$) \\
\quad Dropout & $\{0.0, 0.2, 0.4, 0.6\}$ & Categorical \\
\quad Batch norm & $\{\text{True}, \text{False}\}$ & Categorical \\[6pt]
\multicolumn{3}{l}{\textit{Training}} \\[2pt]
\quad Number of epochs & $\{100, 300, 500\}$ & Categorical \\
\quad Batch size & $[16, 128]$ & Discrete ($\log_2$) \\
\quad Weight decay & $[10^{-6}, 10^{-1}]$ & Loguniform \\
\quad Base learning rate & $[10^{-5}, 10^{-2}]$ & Loguniform \\
\bottomrule
\end{tabular}
\end{center}
\end{table}

\subsection{Gated Recurrent Unit (GRU)}

The GRU baseline operates on a binned raster, projecting spike counts at each timestep into a fixed hidden dimension via a linear layer before processing the sequence through a multi-layer GRU. The context window is processed 
in strides determined by the bin size, rather than ingested all at once, reflecting the sequential nature of the recurrent architecture. Note that dropout is disabled within the GRU when depth is $1$, as it is only meaningful across stacked recurrent layers. We include bidirectionality as a tunable hyperparameter: when enabled, separate forward and backward passes are concatenated, allowing the model to exploit future context symmetrically with the past. Adaptive average pooling is applied over the temporal dimension before the final linear readout to support both sequence- and frame-level tasks. The upper bound on hidden dimension is reduced from $512$ to $256$ at a bin size of $1$\,ms to satisfy memory constraints, and the learning rate upper bound is reduced to $10^{-3}$ on the licking rate task to mitigate  training instability. We detail the tuning parameters for GRU decoding baseline in Table~\ref{tab:hp_gru}.

\begin{table}[h]
\caption{Hyperparameter search space for the GRU baseline, tuned 
for 100 trials.}
\label{tab:hp_gru}
\begin{center}
\begin{tabular}{l l l}
\toprule
\textbf{Hyperparameter} & \textbf{Search Space} & \textbf{Sampling} \\
\midrule
\multicolumn{3}{l}{\textit{Model}} \\[2pt]
\quad Bin size (ms) & $\{1, 5, 10, 20, 40\}$ & Categorical \\
\quad Depth & $\{1, 2, 3\}$ & Categorical \\
\quad Hidden dim & $[32, 512]$ & Discrete ($\log_2$) \\
\quad Dropout & $\{0.0, 0.2, 0.4, 0.6\}$ & Categorical \\
\quad Bidirectional & $\{\text{True}, \text{False}\}$ & Categorical \\[6pt]
\multicolumn{3}{l}{\textit{Training}} \\[2pt]
\quad Number of epochs & $\{100, 300, 500\}$ & Categorical \\
\quad Batch size & $[16, 128]^\dagger$ & Discrete ($\log_2$) \\
\quad Weight decay & $[10^{-6}, 10^{-1}]$ & Loguniform \\
\quad Base learning rate & $[10^{-5}, 10^{-2}]$ & Loguniform \\
\bottomrule
\end{tabular}
\end{center}
{\small $^\dagger$ Batch size and hidden dim are restricted to 
$[16, 64]$ and $[32, 256]$ respectively when bin size is 1ms.}
\end{table}

\subsection{POYO}
POYO operates directly on spike events without binning, using the spike tokenization described in Appendix~\ref{app:neural_encoder}.
We detail the tuning parameters for POYO-SS baseline in Table~\ref{tab:hp_poyo}.

\begin{table}[h]
\caption{Hyperparameter search space for the POYO baseline, tuned 
for 100 trials.}
\label{tab:hp_poyo}
\begin{center}
\begin{tabular}{l l l}
\toprule
\textbf{Hyperparameter} & \textbf{Search Space} & \textbf{Sampling} \\
\midrule
\multicolumn{3}{l}{\textit{Model}} \\[2pt]
\quad Embedding dim & $\{64, 96, 128, 192, 256\}$ & Categorical \\
\quad Number of heads & $\{1, 2, 4, 8\}$ & Discrete ($\log_2$) \\
\quad Depth & $\{0, 1, \ldots, 8\}$ & Discrete \\
\quad Dropout & $[0.0, 0.9]$ & Discrete (step 0.1) \\[6pt]
\multicolumn{3}{l}{\textit{Training}} \\[2pt]
\quad Number of epochs & $[100, 1000]$ & Discrete (step 100) \\
\quad Batch size & $[8, 128]$ & Discrete ($\log_2$) \\
\quad Weight decay & $[10^{-3}, 10^{-1}]$ & Discrete ($\log_{10}$) \\
\quad Base learning rate & $[10^{-5}, 10^{-2}]$ & Loguniform \\
\bottomrule
\end{tabular}
\end{center}
\end{table}

\subsection{NDT*}
NDT* operates on binned spike counts using the bin tokenization described 
in Appendix~\ref{app:neural_encoder}. 
We detail the tuning parameters for NDT*-SS baseline in Table~\ref{tab:hp_ndt}.

\begin{table}[h]
\caption{Hyperparameter search space for the NDT* baseline, tuned 
for 100 trials.}
\label{tab:hp_ndt}
\begin{center}
\begin{tabular}{l l l}
\toprule
\textbf{Hyperparameter} & \textbf{Search Space} & \textbf{Sampling} \\
\midrule
\multicolumn{3}{l}{\textit{Model}} \\[2pt]
\quad Embedding dim & $\{64, 96, 128, 192, 256\}$ & Categorical \\
\quad Number of heads & $\{1, 2, 4, 8\}$ & Discrete ($\log_2$) \\
\quad Depth & $\{0, 1, \ldots, 8\}$ & Discrete \\
\quad Dropout & $[0.0, 0.9]$ & Discrete (step 0.1) \\
\quad Bin size & $\{1, 5, 10, 20, 40\}$ ms & Categorical \\[6pt]
\multicolumn{3}{l}{\textit{Training}} \\[2pt]
\quad Number of epochs & $[100, 1000]$ & Discrete (step 100) \\
\quad Batch size & $[8, 128]$ & Discrete ($\log_2$) \\
\quad Weight decay & $[10^{-3}, 10^{-1}]$ & Discrete ($\log_{10}$) \\
\quad Base learning rate & $[10^{-5}, 10^{-2}]$ & Loguniform \\
\bottomrule
\end{tabular}
\end{center}
\end{table}

\subsection{Computational cost of single-session tuning}
\label{app:ss_compute}

The single-session baselines require a separate hyperparameter search for every held-out session. We use a 100-trial Optuna search for each of the 23 evaluation sessions, resulting in 2{,}300 training runs per model. In contrast, \ours\ and the multi-session backbones use a fixed hyperparameter configuration across sessions and require only a single fine-tuning run for each new session. Table~\ref{tab:ss_compute} compares the resulting computational cost.

\begin{table}[h]
\centering
\caption{\footnotesize{\textbf{Estimated computational cost.} For single-session baselines, run times are reported as mean $\pm$ SD across tuning runs, with the median in parentheses, and total GPU-hours are estimated over all 2{,}300 runs. For pretrained models, we report the one-time pretraining cost and the cost of fine-tuning across the 23 evaluation sessions.}}
\label{tab:ss_compute}
\footnotesize
\begin{tabular}{lccc}
\toprule
\textbf{Stage} & \textbf{Runs} & \textbf{GPU-time / run} & \textbf{Total GPU-hours} \\
\midrule
\multicolumn{4}{l}{\textit{Single-session tuning (per model)}} \\[2pt]
\quad Linear   & 2{,}300 & $59.9 \pm 102.7$ min (22.3) & $\sim$2{,}300 \\
\quad MLP      & 2{,}300 & $53.7 \pm 122.5$ min (20.8) & $\sim$2{,}100 \\
\quad GRU      & 2{,}300 & $46.8 \pm 80.9$ min (16.9)  & $\sim$1{,}800 \\
\quad POYO-SS  & 2{,}300 & $55.8 \pm 142.4$ min (17.0) & $\sim$2{,}100 \\
\quad NDT*-SS  & 2{,}300 & $23.6 \pm 19.2$ min (19.0)  & $\sim$900 \\
\quad \textit{Total} & 11{,}500 & -- & $\sim$9{,}200 \\
\midrule
\multicolumn{4}{l}{\textit{Pretrained models (\ours\ and multi-session backbones)}} \\[2pt]
\quad Pretraining (one-time) & 1 & $\sim$12 h & $\sim$12 \\
\quad Fine-tuning & 23 & $\sim$10 min & $\sim$4 \\
\quad \textit{Total} & 24 & -- & $\sim$16 \\
\bottomrule
\end{tabular}
\end{table}

Across the five single-session baselines, the session-specific searches require approximately 9{,}200 GPU-hours in total, with individual models requiring roughly 900–2{,}300 GPU-hours. By comparison, fine-tuning a pretrained model takes about $10 \pm 14$ minutes per session on a single GPU (median $\approx 8$ minutes), or approximately 4 GPU-hours across all 23 evaluation sessions. The pretraining cost is incurred once and can then be shared across target sessions, whereas the tuning cost of the single-session baselines must be repeated for each new session.

\paragraph{Amortized cost.}
While the comparison between single-session and pretrained models is not the focus of this work, it is important for interpreting the results in the context of prior neural decoding studies. In our experiments, each single-session baseline is extensively tuned for every held-out recording, allowing even relatively simple models to achieve strong performance. Earlier neural decoding studies may not have used the same amount of hyperparameter tuning for every target session~\cite{azabou2024unified, ryoo2025generalizable}, which may partly explain why our single-session baselines are stronger than those previously reported and can, in some cases, outperform pretrained models that were previously shown to improve over session-specific models. Recent benchmarks have similarly highlighted the strength of carefully tuned single-session models~\cite{andre2026brainwidebench}.

This is consistent with a broader motivation for pretrained and generalist models~\cite{kaiser2017one, reed2022generalist, lachi2024graphfm, hollmann2025accurate}. The benefit of such models is not necessarily to outperform a separately optimized specialist on every target dataset, but to provide strong performance with limited target-specific tuning. In our setting, the pretrained models use a fixed hyperparameter configuration across sessions and require only a short fine-tuning stage. The upfront cost of pretraining can therefore be amortized across sessions, animals, and tasks, avoiding the need to repeat an expensive hyperparameter search for each new recording.
\section{Additional Results}




\begin{figure*}[t]
  \centering
  \includegraphics[width=\linewidth]{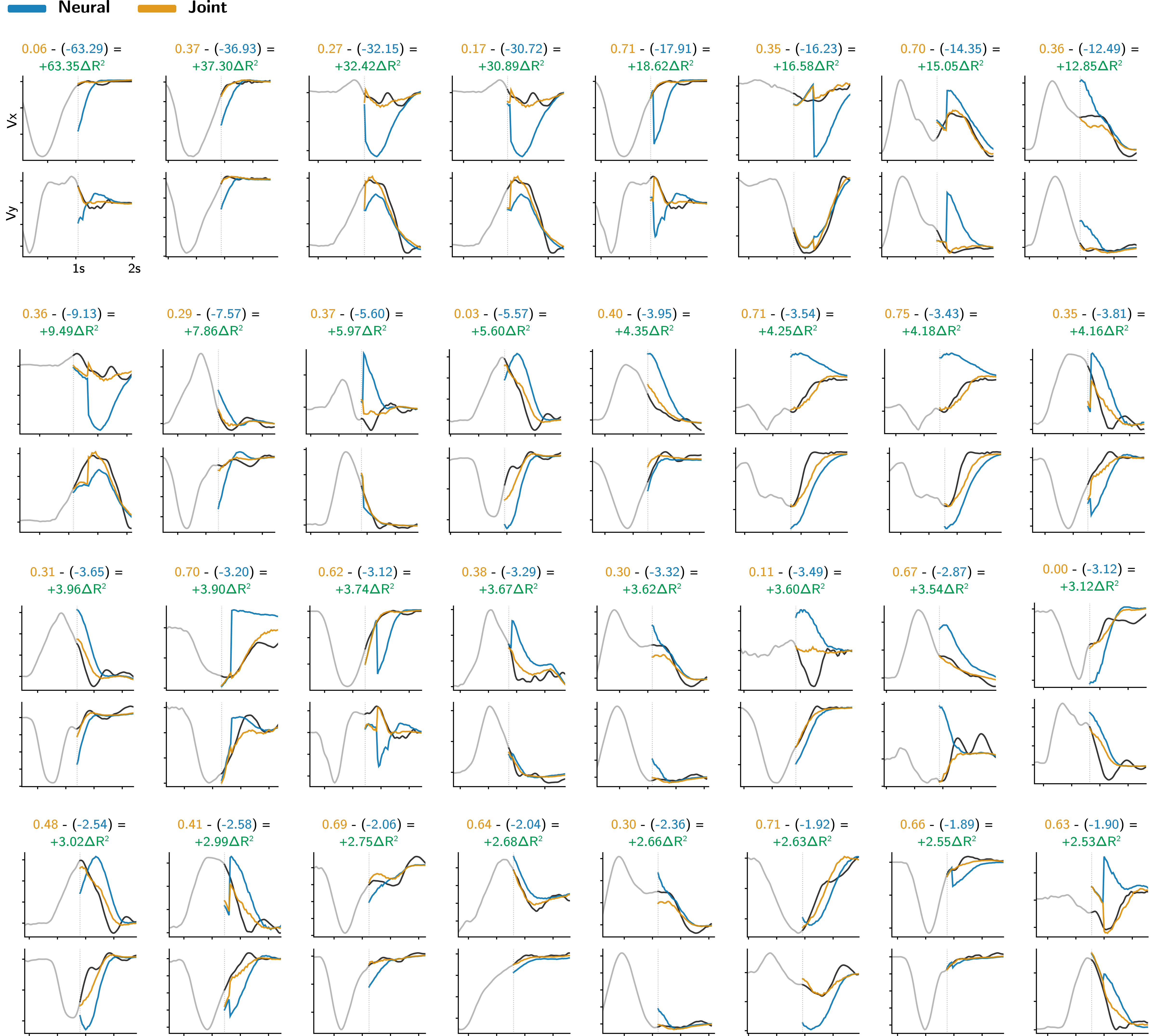}
\caption{\footnotesize{\textbf{Behavioral context improves neural decoding.}  Predicted velocity trajectories ($v_x$, $v_y$) under neural-only (blue) and 
joint neural-behavior (orange) context, vs. ground truth (black),
across 1~sec decoding windows sorted by $\Delta R^2$ 
(largest to smallest improvement). The gray trace shows the preceding behavioral context provided to the joint model. $R^2$ values are reported as 
$R^2_{\text{joint}} - R^2_{\text{neural}} = \Delta R^2$. Behavioral context consistently produces smoother, better-calibrated trajectories, with the largest gains occurring when neural-only decoding fails to constrain trajectory amplitude and timing}}
  \label{fig:additional_trajectories}
\end{figure*}

\subsection{Session-wise statistical analysis}
\label{app:sessionwise_stats}

\begin{table}[h]
\centering
\caption{\footnotesize{\textbf{Session-wise paired comparisons of \ours\ and matched backbones.} Values report mean $R^2$ $\pm$ standard deviation across held-out sessions (each session averaged over three seeds). Paired $t$-tests (two-sided) compare each \ours\ variant with its matched multi-session backbone and were conducted separately for each task ($n=6$ for CO and RT, $n=11$ for IW). Significant results ($p<0.05$) are bolded.}}
\label{tab:sessionwise_ttest}
\resizebox{0.85\textwidth}{!}{
\begin{tabular}{llcccc}
\toprule
\textbf{Backbone} & \textbf{Task} & \textbf{\ours} & \textbf{Baseline} & \textbf{Paired $\boldsymbol{\Delta R^2}$} & \textbf{Paired $t$-test ($\boldsymbol{p}$)} \\
\midrule
NDT* & CO & $0.9027 \pm 0.0138$ & $0.9010 \pm 0.0175$ & $+0.0017$ & $0.4240$ \\
     & RT & $0.7734 \pm 0.0716$ & $0.7600 \pm 0.0789$ & $\mathbf{+0.0134}$ & $\mathbf{0.0123}$ \\
     & IW & $0.7802 \pm 0.0460$ & $0.7767 \pm 0.0435$ & $+0.0035$ & $0.1920$ \\
\midrule
POYO & CO & $0.9031 \pm 0.0178$ & $0.8973 \pm 0.0185$ & $+0.0058$ & $0.1875$ \\
     & RT & $0.7609 \pm 0.0726$ & $0.7498 \pm 0.0742$ & $\mathbf{+0.0112}$ & $\mathbf{0.0001}$ \\
     & IW & $0.7361 \pm 0.0644$ & $0.7237 \pm 0.0680$ & $\mathbf{+0.0124}$ & $\mathbf{0.0089}$ \\
\bottomrule
\end{tabular}
}
\end{table}

Table~\ref{tab:boyo_asymmetric} reports mean performance $\pm$ standard error over three training seeds. This captures variability across training runs, but does not reflect how performance varies across held-out recording sessions. We therefore also report the corresponding session-wise statistics in Table~\ref{tab:sessionwise_ttest}. For each session, we compute $R^2$ over all test trials and average across the three seeds. We then compare each \ours\ variant with its matched multi-session backbone using a two-sided paired $t$-test across sessions, separately for each task ($n=6$ for CO and RT, and $n=11$ for IW).

Table~\ref{tab:sessionwise_ttest} reports mean $R^2 \pm$ standard deviation across sessions together with the mean paired difference and exact $p$-value. The session-wise standard deviation captures variation across recording sessions, whereas the standard error in Table~\ref{tab:boyo_asymmetric} captures variability across training runs. \ours-POYO significantly improves over POYO on RT ($p=0.0001$) and IW ($p=0.0089$), and \ours-NDT* significantly improves over NDT* on RT ($p=0.0123$). The improvements on CO are not significant for either backbone ($p=0.1875$ for POYO and $p=0.4240$ for NDT*), where within-task performance is already high. The improvement of \ours-NDT* on IW is also not significant ($p=0.1920$), consistent with the weaker transfer to IW discussed in Section~\ref{sec:task_ablation}.

\subsection{Comparison with Kalman-based decoders}
\label{app:kalman}

We further compare \ours\ against Kalman-based decoders that explicitly model behavioral dynamics. We consider three baselines. First, we fit a standalone Kalman-filter decoder separately for each session (Kalman Filter SS). Second, we apply a session-specific Kalman smoother to POYO’s predictions as a post hoc temporal regularizer (POYO + Kalman Smoother SS). Third, we fit a single Kalman smoother using behavioral trajectories pooled across the CO pretraining sessions and apply it to POYO predictions on both CO and RT (POYO + Kalman Smoother MS).

\begin{table}[h]
\centering
\caption{\footnotesize{\textbf{Kalman-based baselines.} Models pretrained on CO and evaluated on CO and RT. Values are reported as mean $\pm$ SEM across three seeds.}}
\label{tab:kalman}
\resizebox{0.6\linewidth}{!}{
\begin{tabular}{lcc}
\toprule
\textbf{Method} & \textbf{CO} & \textbf{RT} \\
\midrule
Kalman Filter (SS) & $0.4000 \pm 0.0381$ & $0.4066 \pm 0.0399$ \\
POYO + Kalman Smoother (SS) & $0.8982 \pm 0.0062$ & $0.7515 \pm 0.0018$ \\
POYO + Kalman Smoother (MS) & $0.9004 \pm 0.0015$ & $0.7508 \pm 0.0017$ \\
POYO & $0.8973 \pm 0.0013$ & $0.7498 \pm 0.0015$ \\
\ours-POYO & $0.9031 \pm 0.0003$ & $0.7609 \pm 0.0014$ \\
\bottomrule
\end{tabular}
}
\end{table}

As shown in Table~\ref{tab:kalman}, the standalone Kalman decoder performs substantially worse than the learned neural decoders. Applying Kalman smoothing to POYO gives a small improvement over POYO in both the session-specific and pooled settings, indicating that an explicit dynamics prior can be beneficial. However, \ours-POYO achieves the best performance on both CO and RT. The pooled Kalman smoother provides a smaller improvement on RT than the session-specific smoother, likely because its dynamics are learned only from CO behavioral trajectories and then transferred to RT without access to RT behavior.

We report Kalman results only on CO and RT, which come from the same dataset and therefore share the same behavioral sampling rate. IW is sampled differently, so the same Kalman dynamics model cannot be applied directly and would need to be refit. In contrast, \ours\ can incorporate behavioral data across sessions and tasks with different sampling rates without requiring a separate task-specific dynamics model.

\subsection{Task-specific behavioral data on RT}
\label{app:rt_behavior}

To test whether the benefit of task-matched behavioral pretraining observed on IW (Section~\ref{sec:task_ablation}) extends to RT, we repeat the experiment using \ours-POYO. Starting from a model pretrained without unpaired behavioral data, we first add 6.5 hours of unpaired CO behavioral data and then an additional 2 hours of unpaired RT behavioral data. As shown in Table~\ref{tab:rt_behavior}, CO behavioral data improves RT decoding, and adding task-matched RT behavioral data provides a further gain, consistent with the trend observed on IW.

\begin{table}[h]
\centering
\caption{\footnotesize{\textbf{Task-specific behavioral data on RT.} \ours-POYO evaluated on held-out RT sessions with increasing amounts of unpaired behavioral data during pretraining (single seed).}}
\label{tab:rt_behavior}
\footnotesize
\begin{tabular}{lc}
\toprule
\textbf{Unpaired behavioral data} & \textbf{RT} \\
\midrule
None & 0.7511 \\
+ 6.5 h CO & 0.7637 \\
+ 6.5 h CO + 2 h RT & 0.7705 \\
\bottomrule
\end{tabular}
\end{table}

\subsection{Ablation results for CO task}

Table~\ref{tab:ablations_full} extends the ablation results from Section~\ref{sec:ablation} to include within-task evaluation on the CO task in addition to the cross-task results reported in Section~\ref{sec:ablation}. The trends observed in the Section~\ref{sec:ablation} hold consistently across all tasks: removing behavior dropout leads to significant performance degradation below even the base model baselines, and replacing the decoder with \ours's deeper architecture without access to unpaired behavioral data does not recover \ours's performance. Together, these results confirm that the gains from \ours\ are attributable to the asymmetric training strategy rather than architectural differences.
\begin{table}[h]
\centering
\caption{\footnotesize{\textbf{Ablation study:}
Values are reported as mean $\pm$ SEM across three seeds.}}
\label{tab:ablations_full}
\vspace{1mm}
\resizebox{0.8\linewidth}{!}{
\begin{tabular}{l|c|cc}
\toprule
& \textbf{Within-Task} & \multicolumn{2}{c}{\textbf{Cross-Task}} \\
\cmidrule(lr){2-2} \cmidrule(lr){3-4}
\textbf{Model Variant} & \textbf{CO} & \textbf{RT} & \textbf{IW} \\
\midrule
\textbf{\ours-POYO} & $0.9031 \pm 0.0003$ & $0.7609 \pm 0.0014$ & $0.7361 \pm 0.0020$ \\
\quad w/o dropout & $0.8179 \pm 0.0035$ & $0.6925 \pm 0.0133$ & $0.6898 \pm 0.0035$ \\
\quad w/o unpaired & $0.8992 \pm 0.0012 $ & $0.7476 \pm 0.0012 $ & $0.6849 \pm 0.0017$ \\
\midrule
\textbf{\ours-NDT} & $0.9027 \pm 0.0016$ & $0.7734 \pm 0.0018$ & $0.7802 \pm 0.0001$ \\
\quad w/o dropout & $0.8110 \pm 0.0062$ & $0.7101 \pm 0.0083$ & $0.7266 \pm 0.0019$ \\
\quad w/o unpaired & $0.9018 \pm 0.0009$ & $0.7683 \pm 0.0030$ & $0.7757 \pm 0.0011$ \\
\bottomrule
\end{tabular}
}
\end{table}

\subsection{Additional Trajectory Examples}
\label{app:additional_trajectories}

Figure~\ref{fig:additional_trajectories} extends the trajectory visualizations from Section~\ref{sec:results}, showing additional 1-second decoding windows sorted by $\Delta R^2$ (largest to smallest improvement from joint over neural-only decoding). The gray trace shows the preceding behavioral context provided to the joint model. Consistent with Section~\ref{sec:traj}, behavioral context produces smoother and better-calibrated trajectories across a broad range of decoding windows, with the largest gains occurring when neural-only decoding fails to constrain trajectory amplitude and timing.

\clearpage

\end{document}